# Sentence Alignment with Parallel Documents Facilitates Biomedical Machine Translation

Shengxuan Luo[1,2], Huaiyuan Ying[1,2], Jiao Li[3], Sheng Yu[1,2]

[1]*Center for Statistical Science, Tsinghua University, Beijing, China, Beijing, China;*

[2]*Department of Industrial Engineering, Tsinghua University, Beijing, China;*

[3]*Institute of Medical Information/Library, Chinese Academy of Medical Sciences/Peking Union Medical College, Beijing, China.*

**Correspondence to**:

Sheng Yu

Weiqinglou Rm 209, Tsinghua University

Beijing, 100084, China

Email: syu@tsinghua.edu.cn

Tel: +86-10-62783842

## ABSTRACT

**Objective**: Today's neural machine translation (NMT) can achieve near human-level translation quality and greatly facilitates international communications, but the lack of parallel corpora poses a key problem to the development of translation systems for highly specialized domains, such as biomedicine. This work presents an unsupervised algorithm for deriving parallel corpora from document-level translations by using sentence alignment and explores how training materials affect the performance of biomedical NMT systems.

**Materials and Methods**: Document-level translations are mixed to train bilingual word embeddings (BWEs) for the evaluation of cross-lingual word similarity, and sentence distance is defined by combining semantic and positional similarities of the sentences. The alignment of sentences is formulated as an extended earth mover's distance problem. A Chinese-English biomedical parallel corpus is derived with the proposed algorithm using bilingual articles from UpToDate and translations of PubMed abstracts, which is then used for the training and evaluation of NMT.

**Results**: On two manually aligned translation datasets, the proposed algorithm achieved accurate sentence alignment in the 1-to-1 cases and outperformed competing algorithms in the many-to-many cases. The NMT model fine-tuned on biomedical data significantly improved the in-domain translation quality (zh-en: +17.72 BLEU; en-zh: +17.02 BLEU). Both the size of the training data and the combination of different corpora can significantly affect the model's performance.

**Conclusion**: The proposed algorithm relaxes the assumption for sentence alignment and effectively generates accurate translation pairs that facilitate training high quality biomedical NMT models.

# 1 INTRODUCTION

In recent years, neural machine translation (NMT) systems have achieved near human-level performance on several language pairs in the general domain.[1-3] However, the reliance on large parallel corpora is the major bottleneck of training an NMT system, and in many domains, the vast majority of language pairs have very little, if any, parallel data.[4, 5] Manually obtaining such parallel corpora is expensive and labor intensive.

In the biomedical domain, machine translation is essential to facilitate international collaborations. The outbreak of the COVID-19 pandemic showed the importance of international medical communication and cross-language medical data sharing in helping humans around the world respond to public health emergencies.[6, 7] A biomedical machine translation system can facilitate the exchange of treatment experiences and research findings in a timely and efficient manner in countries with different languages. In addition to communication, biomedical neural machine translation (NMT) is needed for many other applications, such as adapting statistical or machine learning models developed in one country to countries in which other languages are spoken or unifying the terminology systems and ontologies of different countries for global collaboration in medical knowledge and artificial intelligence.

The architecture of machine translation systems has developed over a long period of time. Statistical machine translation used to be the mainstream machine translation,[8-10] while NMT has become the de facto standard for large-scale machine translation in the past few years.[11-14] Currently, the most popular machine translation model is the Transformer.[13]

The first and most crucial step in constructing a translation model is to obtain a parallel corpus since the essence of machine translation (MT) is to use the patterns mined from existing corpora to translate new text.[15] Most current work follows a pipeline to construct a parallel corpus: text collection, text preprocessing, and sentence alignment.[15-18] In the public domain of some dominant languages, there are many large-scale publicly available parallel corpora containing tens of millions of sentence pairs. In contrast, parallel corpora in the biomedical domain are significantly smaller and less available. On the other hand, the biomedical domain is well known for its complex nomenclature,[19] massive amounts of terminologies, and a language style that differs from that of the general domain. As a result, translation models trained in the public domain perform poorly in the biomedical domain [20, 21] and lead to lower translation quality. The biomedical NMT model needs to learn how to translate biomedical texts precisely from domain corpora. Some existing translation works have provided bilingual biomedical corpora for a few major languages, including de/en, es/en, fr/en, it/en, pt/en, and zh/en[1]. Their data size ranges from thousands to hundreds of thousands of sentence pairs,[19, 22, 23] which is insufficient to train NMT models.

Noting that there are document-level translations (parallel documents) available from research papers and multilanguage websites, in this paper, we propose a novel unsupervised sentence alignment method to extract biomedical sentence pairs from parallel documents to develop a high-quality biomedical NMT model. The method first mixes the paired documents of two languages into a pseudodocument according to the relative position of the words in the documents. These pseudodocuments are then used to train bilingual word embeddings (BWEs) to evaluate bilingual word similarities.[24] To align the sentences of a parallel document pair, we define sentence distance by combining the word similarity and the relative position of the sentences in the document. Finally, the alignment of sentences is formulated as an extended earth mover's distance (EMD) optimization problem to transfer the information from the source language to the target language. To establish a biomedical parallel corpus for English-Chinese machine translation, we apply the proposed sentence alignment method to the bilingual articles from UpToDate [25] and translations of abstracts from PubMed. We also explore various settings for training biomedical NMT models by mixing general domain and biomedical corpora and evaluate their performance and generalizability.

## 2 RELATED WORK

---

[1] de= German, en=English, es=Spanish, fr=French, it=Italian, pt=Portuguese, zh=Chinese.

This section reviews related works involving BWEs, sentence alignment tasks, and the EMD problem.

**2.1 Bilingual word embeddings**

BWEs aim to produce a common vector space for words of two languages. BWEs usually maintain the semantic information of both languages, such as the word similarity that is usually computed by the cosine between their embedding vectors.[21]

Current approaches to constructing BWEs can be classified into two categories: (1) methods that train embeddings for two languages separately and learn a mapping to align the two embedding spaces with bilingual information [26-28] or unsupervised approaches [29, 30] and (2) methods that train BWEs by learning a shared embedding space for both languages.[24, 31] Our method for BWE is similar to Vulić et al.[24] in that aligned documents of two languages are merged into one document to train word embeddings. Since our data comprise whole-document translations, the positions of the words are also utilized when merging the documents to generate better BWEs, instead of using shuffling as in Vulić et al.[24]

**2.2 Sentence alignment**

Sentence alignment refers to the task of aligning sentences in a document pair. The aligned sentence pairs express the same meaning in different languages. The most common alignment is 1-to-1 alignment. However, there is a significant presence of complex alignment relationships, including 1-to-0, 0-to-1 and many-to-many cases, such as 1-to-2, 1-to-3, and 2-to-2, due to the characteristics of languages, translators' personal reasons, or errors in sentence segmentation. Most previous work on sentence alignment can be classified into three categories: length-based,[32, 33] word-based,[34, 35] and translation-based.[16, 36] Some of these models are supervised and depend on dictionaries or existing sentence pairs. Furthermore, strong assumptions about the alignment types are commonly seen in these models, e.g., assuming that there are only specific types of alignment or the alignment must be ordinal. Most models are weak in the many-to-many case. We propose a novel unsupervised sentence alignment method that achieved high accuracy in both 1-to-1 and many-to-many cases. Neither external information nor unnecessary restriction on the form of the alignment is required.

**2.3 Earth mover's distance**

EMD is the optimization problem of minimizing the cost of moving all mounds into holes, where the volumes of the mounds and holes are known. The mounds and holes have equal total volumes. The cost of moving the earth is the product of the moving distance and the moving amount. The EMD problem is solved as a linear program (see more details in Section 3.3), and the solution gives a transport matrix (the volume from each mound to each hole). EMD has been used in the bilingual lexicon induction task.[27, 37, 38] Given the weight of words and the dissimilarity between words, the transport matrix obtained by solving the EMD infers the translation relation of words between the two languages. Our method is inspired by these works and applies an extended EMD to the sentence alignment scenario.

# 3 Methods

This section introduces unsupervised sentence alignment using parallel documents for building a biomedical machine translation system. Figure 1 illustrates the 4 steps of the process, namely, BWE from parallel documents, defining sentence distance, sentence alignment as an EMD optimization problem, and translation model training.

**3.1 BWE from parallel documents**

Parallel documents can be used to develop BWEs to measure word similarity across two languages. As translations are generally sentences for sentences, it is reasonable to assume that bilingual word pairs usually appear in similar positions in parallel documents. Based on this assumption, we mix two parallel documents into one document by reranking the words according to their relative positions in their documents. If $S$ and $T$ are parallel documents including $N$ words and $M$ words, respectively, the relative

6both 1-to-1 and many-to-many cases. Neither external information nor unnecessary restriction on the form of the alignment is required.

**2.3 Earth mover's distance**

EMD is the optimization problem of minimizing the cost of moving all mounds into holes, where the volumes of the mounds and holes are known. The mounds and holes have equal total volumes. The cost of moving the earth is the product of the moving distance and the moving amount. The EMD problem is solved as a linear program (see more details in Section 3.3), and the solution gives a transport matrix (the volume from each mound to each hole). EMD has been used in the bilingual lexicon induction task.[27, 37, 38] Given the weight of words and the dissimilarity between words, the transport matrix obtained by solving the EMD infers the translation relation of words between the two languages. Our method is inspired by these works and applies an extended EMD to the sentence alignment scenario.

# 3 Methods

This section introduces unsupervised sentence alignment using parallel documents for building a biomedical machine translation system. Figure 1 illustrates the 4 steps of the process, namely, BWE from parallel documents, defining sentence distance, sentence alignment as an EMD optimization problem, and translation model training.

**3.1 BWE from parallel documents**

Parallel documents can be used to develop BWEs to measure word similarity across two languages. As translations are generally sentences for sentences, it is reasonable to assume that bilingual word pairs usually appear in similar positions in parallel documents. Based on this assumption, we mix two parallel documents into one document by reranking the words according to their relative positions in their documents. If $S$ and $T$ are parallel documents including $N$ words and $M$ words, respectively, the relative position of the $i$-th word in $S$ is $\frac{i}{N}$, and the relative position of the $j$-th word in $T$ is $\frac{j}{M}$. We remove sentence boundaries and rearrange all the words in the parallel documents in ascending order by their relative positions to generate new bilingual pseudodocuments, from which BWEs are trained with the word2vec [26] Skip-gram model. The similarity between two words is defined as the cosine of their embedding vectors in BWEs.

**3.2 Sentence distance**

The distance between two sentences of parallel documents can be defined by the words' similarity and the sentences' relative positions in the documents. Assume $S = \{s_i\}_{i=1}^n$, $T = \{t_j\}_{j=1}^m$, where $s_i$ and $t_j$ denote the sentences in the parallel documents $S$ and $T$.

The word-based distance $d_1(s_i, t_j)$ is defined as

$$d_1(s_i, t_j) = \left( \frac{\sum_{w_k \in s_i} \max_{v_l \in t_j} \cos(w_k, v_l)}{len(s_i)} \right)^{-1},$$

where $w_k$ and $v_l$ are words in $s_i$ and $t_j$, respectively. $d_1(s_i, t_j)$ searches the most similar word in $t_j$ for each word in $s_i$ and averages the similarity to measure the word-based distance between sentences.

The distance based on the relative position is defined as

$$d_2(s_i, t_j) = |posi(s_i) - posi(t_j)|^3,$$

where

$$posi(s_i) = \frac{\sum_{p<i} len(s_p)}{N},$$

$$posi(t_j) = \frac{\sum_{p<j} len(t_p)}{M}.$$

The definitions of $N$ and $M$ are the same as in Section 3.1.

Since the relative positions of aligned sentences are not exactly the same, the positional distance $d_2$ should be small when the relative positions of the sentences are close. Therefore, the positional distance $d_2$ is defined as the third power of the difference of relative positions.

Finally, the distance between $s_i$ and $t_j$ can be defined as

$$D_{ij} = d_1(s_i, t_j) + \alpha d_2(s_i, t_j).$$

In practice, we find $\alpha = 1$ generally works well.

### 3.3 Sentence alignment

To formulate sentence alignment as an optimization problem, we temporarily assume that in a pair of parallel documents, $(S, T)$, $S$ and $T$ contain an equal amount of information, where the amount of information in one sentence is proportional to the sentence's relative length. To see sentence alignment as an EMD, information is treated as earth, and we treat the sentences in $S$ as mounds and the sentences in $T$ as holes and take the distance from the mound to the hole as the sentence distance defined in Section 3.2. Furthermore, let the volume of the mound $s_i$ and hole $t_j$ be $len(s_i)$ and $len(t_j)$, respectively, where

$$len(s_i) = \frac{\#\{w_k | w_k \in s_i\}}{N},$$

$$len(t_j) = \frac{\#\{w_k | w_k \in t_j\}}{M}.$$

$D_{ij}$ is the distance between sentences $s_i$ and $t_j$. Then, the problem of sentence alignment can be formulated as solving the EMD from $S$ to $T$:

$$\min_P \sum_{i,j} D_{ij} P_{ij},$$

$$\text{s.t.} \sum_j P_{ij} \leq len(s_i), \forall i \in \{1,2,\dots,n\},$$

$$\sum_i P_{ij} = len(t_j), \forall j \in \{1,2,\dots,m\},$$

$$P_{ij} \geq 0, \forall i,j,$$

where $P$ is the transport matrix and $P_{ij}$ denotes the transport volume from $s_i$ to $t_j$. The first constraint guarantees that the amount of earth moved from each mound does not exceed the volume of the mound, while the second constraint means that the information is completely transported. Since $\sum_{i,j} P_{ij} = \sum_j len(t_j) = 1 = \sum_i len(s_i)$, the "≤" in the first constraint is effectively equivalent to "=".

However, in real data, the above strict constraints result in elements in $P$ corresponding to some unaligned sentences being nonzero since the amount of information contained in the sentence is not strictly proportional to sentence length. These nonzero elements in $P$ lead to incorrect many-to-many sentence alignment. Therefore, we introduce a relaxation factor to transform the problem into the following:

$$\min_P \sum_{i,j} D_{ij} P_{ij},$$

$$\text{s.t.} \sum_j P_{ij} \leq len(s_i) + \frac{\varepsilon}{n}, \forall i \in \{1, 2, \dots, n\},$$

$$\sum_i P_{ij} \leq len(t_j) + \frac{\varepsilon}{m}, \forall j \in \{1, 2, \dots, m\},$$

$$P_{ij} \geq 0, \forall i, j,$$

$$\sum_{i,j} P_{ij} = 1,$$

where $\varepsilon \geq 0$ is the relaxation factor, which can effectively remove nonzero elements that should not appear in $P$. The selection of $\varepsilon$ value is a new problem. It is easy to see that $\varepsilon = 0$ reduces the problem to the previous one. Increasing $\varepsilon$ tends to reduce the number of nonzero elements, and with sufficiently large $\varepsilon$, transport will only occur in one pair of sentences with the smallest distance. To address this problem, we propose a practical selection approach. Since having too many completely nonzero $2 \times 2$ submatrices in $P$ is an indicator of incorrect many-to-many alignments, we define $Z_\varepsilon(P)$ as the sum of the smallest elements in all the completely nonzero $2 \times 2$ submatrices and grid search $\varepsilon$ to minimize $Z_\varepsilon(P) + \gamma\varepsilon$, resulting in a small $\varepsilon$ that can effectively reduce the alignment error. In our experiments, $\gamma = 1$ is a good choice.

Each nonzero element in the solution of $P$ implies an alignment between sentences. In real data, the sentence alignment method occasionally bundles several alignments into one alignment group. We split the alignment groups containing at least three sentences in both languages by the length-based approach, the Gale-Church algorithm,[33] to obtain the final alignments.

### 3.4 Biomedical NMT model training

We followed the traditional method of domain adaptation for NMT to pretrain the NMT model on general corpora and then fine-tuned it with in-domain parallel corpora. We use the base Transformer model as our NMT model.[39][2]

# 4 RESULTS

### 4.1 Data collection

We collected 7,137 pairs of documents in both Chinese and English from the UpToDate website. We also obtained Chinese translations of 60,553 PubMed abstracts contributed by volunteers from a public website[3]. We applied preprocessing for both datasets, including punctuation standardization, sentence boundary detection, truecasing, and Chinese word segmentation.[4]

### 4.2 Quality of BWEs

We used the corpora of UpToDate and PubMed abstracts to train BWEs according to Section 3.1. To test the quality of word similarity evaluated by BWEs, we retrieved all the medical terms in Xiangya Medical Dictionary and randomly selected 200 terms whose English and Chinese forms both appear in a pair of parallel documents and the difference of the relative position for both forms in the parallel documents is less than 0.1. For each selected Chinese term, we manually checked whether the most similar English term found by BWEs was correct. The top 1 and top 10 accuracy in cosine similarity were 66.8% and 86.0%, respectively. The 200 terms are divided into 4 quarters by word frequency in UpToDate (Figure 2). We found that the error was mainly due to inconsistency in the morphological variation and words that are close in meaning. For example, the most similar term in BWEs for "角膜 (cornea)" is "corneal". In addition, frequent terms (in the first quarter) are likely to be common words with many meanings, which leads to low accuracy in this quarter.

---

[2] We implemented the model with the open-source toolkit THUMT.

[3] http://www.chinapubmed.net/.

[4] Chinese word segmentation used the jieba toolkit (https://github.com/fxsjy/jieba) with in-domain lexicons from https://pinyin.sogou.com/dict/.

**4.3 Sentence alignment results**

We compared the proposed method with three sentence alignment methods: the length-based method Gale-Church,[33] the word and length-based method Microsoft Aligner,[34] and the translation-based method Bleualign.[16] We randomly selected parallel documents in UpToDate and PubMed abstracts and manually aligned these documents as test sets. Figure 3 shows the sizes of the test sets. The majority are 1-to-1 alignments, although the proportion of n-to-m alignments with $\min(n, m) \geq 1, \max(n, m) > 1$ is not ignorable. During the manual alignment, we also found that the UpToDate data were better than PubMed abstracts in terms of translation quality. The precision, recall, and F1 scores are shown in Table 1, with the metrics calculated as follows:

$$precision = \frac{\#\{correct\ extracted\ alignments\}}{\#\{all\ extracted\ alignments\}},$$

$$recall = \frac{\#\{correct\ extracted\ alignments\}}{\#\{all\ manual\ alignments\}},$$

$$F_1 = \frac{precision \times recall}{precision + recall}.$$

For the 1-to-1 alignment, our method and Microsoft Aligner are better, and our method achieves the best performance in the n-to-m case for both data. After duplication elimination and cleaning, we ultimately extracted over 1.4 million sentence pairs by applying the proposed method. Specifically, we extracted 788,372 1-to-1 and 344,722 n-to-m sentence pairs from UpToDate and 203,740 1-to-1 and 79,329 n-to-m sentence pairs from PubMed abstracts.

**Table 1.** Precision, recall, and F1 scores of manually aligned sentence pairs. In n-to-m case, $max(n,m) > 1, n, m \neq 0$. GC: Gale-Church; MS: Microsoft Aligner; Bleu: Bleualign; Ours: the proposed method with $\alpha = 1, \gamma = 1$.

|        |      | UpToDate  |       |       | PubMed Abstract |       |       |
|--------|------|-----------|-------|-------|-----------------|-------|-------|
|        |      | Precision | Recall | $F_1$ | Precision | Recall | $F_1$ |
| 1-to-1 | GC   | 69.10 | 74.05 | 71.49 | 89.21 | 78.47 | 83.50 |
|        | MS   | 94.41 | **94.41** | **94.41** | **99.49** | 45.14 | 62.10 |
|        | Bleu | 87.50 | 90.83 | 89.13 | 87.01 | **93.06** | 89.93 |
|        | Ours | **95.59** | 92.17 | 93.85 | 95.76 | 88.89 | **92.20** |
| n-to-m | GC   | 63.20 | 55.63 | 59.18 | 29.80 | 42.45 | 35.01 |
|        | MS   | 88.79 | 72.54 | 79.84 | 75.00 | 2.16 | 4.20 |
|        | Bleu | 79.82 | 61.27 | 69.32 | 44.34 | 33.81 | 38.37 |
|        | Ours | **89.55** | **84.51** | **86.96** | **75.49** | **55.40** | **63.90** |

### 4.4 NMT model results

We established a biomedical domain corpus by merging the resulting 1.4 million sentence pairs from Section 4.3 and the existing parallel corpus, ECCParaCorp,[40] and bilingual terms extracted from a medical dictionary[5]. ECCParaCorp is a Chinese-English parallel corpus of cancer information consisting of 412 pairs of words, 1,083 pairs of phrases, and 5,190 pairs of sentences. The in-domain test set consists of manually aligned sentence pairs (subsection 4.3) and 500 sentence pairs from ECCParaCorp. The in-domain training data are the remaining sentence pairs (denoted as BioMed).

We pretrained the Transformer model with over 24 million Chinese-English sentence pairs from WMT18 for 200,000 steps. Then, the base model was fine-tuned on in-domain training data for 40,000 steps. The batch size was 16,384 tokens in each step. Additionally, a model trained only on in-domain data and a model fine-tuned on a mixed corpus consisting of in-domain and out-of-domain (WMT18) data were used to investigate the effect of pretraining and in-domain tuning (Table 2). Pretraining was necessary because the model benefited in both translation directions by increasing 1.98 BLEU score for zh-en and 1.87 for en-zh. The model fine-tuned on in-domain data significantly improved the in-domain translation quality (zh-en: +17.72 BLEU; en-zh: +17.02 BLEU), while fine-tuning decreased the quality of out-of-domain translation (zh-en: -9.93 BLEU; en-zh: -11.12 BLEU). As shown in Kocmi et al.,[41] a simple mixture of out-of-domain and in-domain data can be harmful for in-domain translation. However, the out-of-domain translation quality improved significantly, while a slight decrease was observed for in-domain data when the model was fine-tuned on the mixture of in-domain and out-of-domain corpora (WMT18/BioMed+WMT18 (small) in Table 2).

---

[5] We extracted 59,532 pairs of terms from a medical dictionary.

**Table 2.** In-domain and out-of-domain performance (BLEU) of NMT models. WMT18/None is the pretrained model. "+" denotes the simple concatenation. WMT18 (small) is a subset of data in WMT18 with the same size as BioMed.

| Pretrain/Fine-tune | In-domain | | newstest2017 | |
|---|---|---|---|---|
| | zh-en | en-zh | zh-en | en-zh |
| WMT18/None | 22.69 | 18.55 | 23.56 | 19.57 |
| None/BioMed | 38.43 | 33.78 | - | - |
| WMT18/BioMed | 40.41 | 35.65 | 13.63 | 8.45 |
| WMT18/BioMed+WMT18 (small) | 40.13 | 35.04 | 22.84 | 18.19 |

We found that even when the domain was limited to biomedicine, the different styles between subdomains were not negligible. To probe this phenomenon, we compared the performance of en-zh models trained with different approaches on a small cancer corpus, ECCParaCorp. The ECCParaCorp covers information on cancer prevention, cancer screening, and cancer treatment, while UpToDate covers comprehensive clinical topics and is much larger than the ECCParaCorp. There are differences in language patterns between the two corpora. Table 3 lists the translation performance on the subdomain data ECCParaCorp from models that were trained by various approaches. It shows that fine-tuning on comprehensive clinical topics (WMT18/UpToDate) does not guarantee good generalizability to a subdomain; indeed, its performance can be inferior to only fine-tuning on the small subdomain data (WMT18/ECCParaCorp). It is also worth noting how training order significantly affects the performance: (1) fine-tuning on the mixture of UpToDate and ECCParaCorp (WMT18/BioMed) did not perform as well as fine-tuning on the target subdomain alone (WMT18/ECCParaCorp); (2) however, fine-tuning first on UpToDate then on ECCParaCorp (WMT18/UpToDate(first)+ECCParaCorp(later)) significantly improved the BLEU score by 10.78. These results suggest that a well-trained biomedical base model can contribute to training a better model in a subdomain, but it is necessary to finish the training on the target subdomain.

**Table 3.** The BLEU score of NMT models in the ECCParaCorp test set in the zh-en direction. The ECCParaCorp data use in training and testing is disjoint.

| Pretrain/Fine-tune | ECCParaCorp test set |
|---|---|
| None/ECC | 7.13 |
| WMT18/None | 11.76 |
| WMT18/UpToDate | 17.30 |
| WMT18/BioMed | 26.34 |
| WMT18/ECCParaCorp | 34.90 |
| WMT18/UpToDate(first)+ECCParaCorp(later) | 45.68 |

## 5 DISCUSSION

As demonstrated by the preceding results, the method we propose can obtain a large number of high-quality 1-to-1 and n-to-m sentence pairs, and the biomedical NMT model has good performance on the BLEU score. In the following section, we discuss the role of the amount of in-domain data and the n-to-m sentence pairs in improving the model performance. In addition, we present examples to explain how the translation quality is improved by fine-tuning and the effects of adding an external in-domain dictionary.

We investigated the relationship between model performance and the number of sentence pairs used in fine-tuning. As Figure 4 shows, to allow the BLEU score to increase linearly would require exponential growth of the size of the fine-tuning data. Moreover, Figure 4 indicates that adding n-to-m sentence pairs improves the model performance consistently under the same number of parallel documents, even though their quality is not as good as 1-to-1 sentence pairs.

A case study is conducted to illustrate the effect of fine-tuning on in-domain data. The example 1 in Figure 5 is in the zh-en direction. The source sentence contains medical terms such as "盐皮质激素",

"螺内酯", "依普利酮", "罗格列酮", and "袢利尿剂". In the fine-tuned model, these terms were correctly translated as "mineralocorticoid", "spironolactone", "eplerenone", "rosiglitazone", and "loop diuretic", whereas the pretrained model generated incorrect translations. The common word "给予" is usually translated to "give", while "administration" is more appropriate in the medical domain, which means "the act of giving a drug to someone". Similar situations are demonstrated in the example 2 in Figure 5. The model learned the translation of "thiazolidinediones" after fine-tuning and translated the term to "噻唑烷二酮类药物" correctly despite decoding different subwords than the target sentence. In terms of language preference, the word "retention" is better translated as the medical term "潴留" instead of the common word "滞留". These examples show that in-domain fine-tuning helps the model learn more terminology and fits the in-domain language style in the decoding phase. Furthermore, in-domain sentence pairs do not cover all biomedical terms. In the example 3 in Figure 5, "aegyptianellosis", "eperythrozoonosis", "grahamellosis", and "haemobartonellosis" never appeared in sentence pairs in the training data (BioMed), and the model was not able to translate them. However, the last three terms appeared in the in-domain dictionary that we added to the training set as a set of bilingual pairs (BioMed*), and the model fine-tuned on BioMed* with the dictionary terms oversampled correctly translated "grahamellosis" and "haemobartonellosis" to "格雷汉体病" and "血巴尔通体病", showing that an additional in-domain dictionary can be helpful.

## 6 CONCLUSION

In this paper, we proposed a new unsupervised sentence alignment method as a linear program that utilizes bilingual word alignment information to evaluate word similarity and further evaluate sentence distance. The proposed method relaxes the assumption about the types of alignment and has better performance on n-to-m alignment. We used all obtained data to build the Chinese-English biomedical translation system. Pretraining the NMT model in the general domain, a larger amount of in-domain data, and n-to-m sentence pairs benefit the biomedical NMT model. In-domain fine-tuning improves the quality of term translation and helps the NMT model fit the in-domain language style in the decoding phase. There are significant differences among subdomains of biomedicine. Based on a well-trained in-domain model, a small amount of data, such as 6,000 sentence pairs, is enough to improve the performance of the NMT model in that subdomain.

**ACKNOWLEDGMENTS**

The authors thank Keming Lu for his help in data collection and processing.

**AUTHOR STATEMENT**

All authors made substantial contributions to the conception and design; acquisition, analysis and interpretation of data; drafting the article or revising it critically for important intellectual content; and final approval of the version to be published.

**CONFLICT OF INTERESTS**: None.

**FUNDING**

This work was supported by the Natural Science Foundation of Beijing Municipality (Grant No. Z190024), the National Natural Science Foundation of China (Grant No. 11801301) , and the National Key R&D Program of China (Grant No. 2016YFC0901901).

**DATA AVAILABILITY STATEMENT**

Source code is made available at: https://github.com/luosx18/biomedical-sentence-aligner

The text materials used for the experiments in this paper can be accessed at: https://www.uptodate.cn/; http://www.chinapubmed.net/; http://www.phoc.org.cn/ECCParaCorp/.

**FIGURE LEGENDS**

**Figure 1.** The workflow of the sentence alignment and model training process.

**Figure 2.** Results split by frequency bins. [0~1/4], [1/4~2/4], [2/4~3/4], and [3/4~4/4] are the four groups of words from high frequency to low frequency.

**Figure 3.** The number of sentence pairs in the test set.

**Figure 4.** The relation between model performance and the size of the in-domain data. At each point, the parallel documents in the 1-to-1 and 1-to-1 & n-to-m cases are the same, but the latter incorporates n-to-m sentence pairs.

**Figure 5.** In example 1 and example 2, the fine-tuning model learns more terminology and fits the in-domain style in zh-en and en-zh. The green words are the correct translation, the red words are incorrect, and the blue word shows the in-domain style of text. In example 3, the effect of mixing the in-domain dictionary and other in-domain data.

**FIGURES**

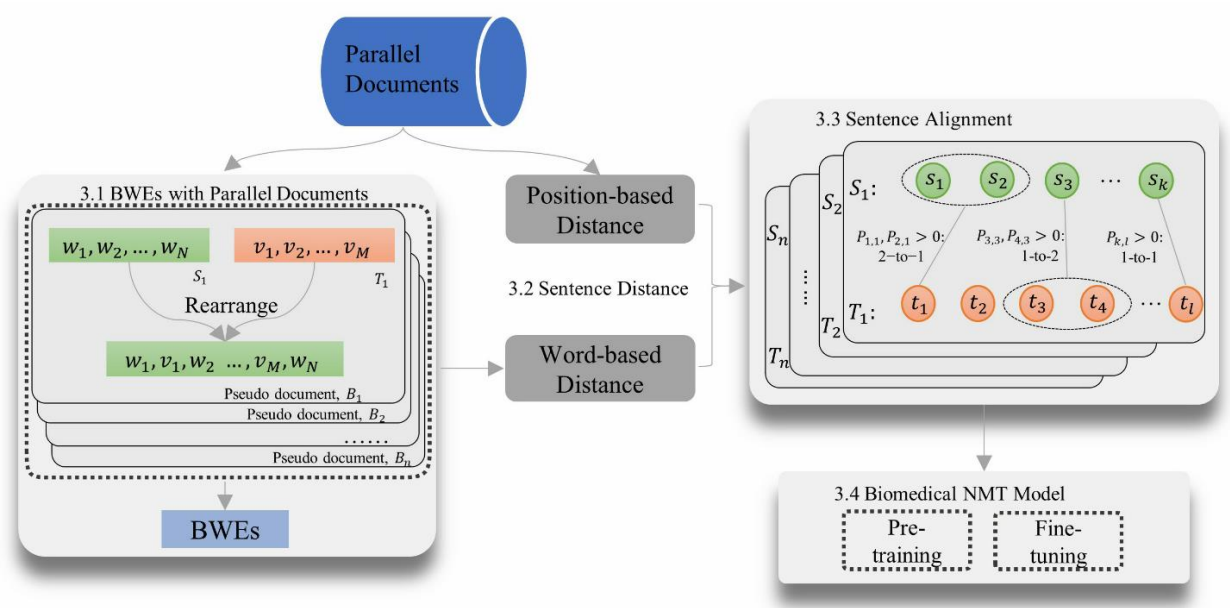

**Figure 1**

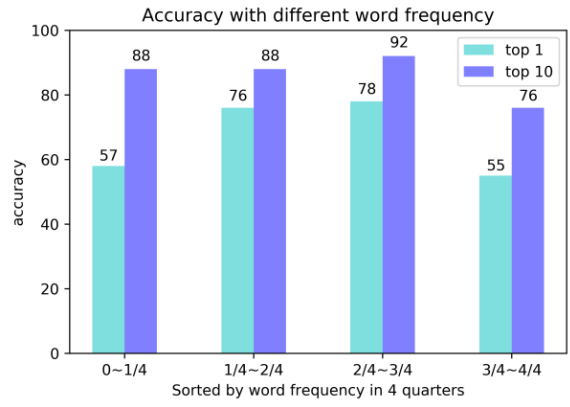

**Figure 2**

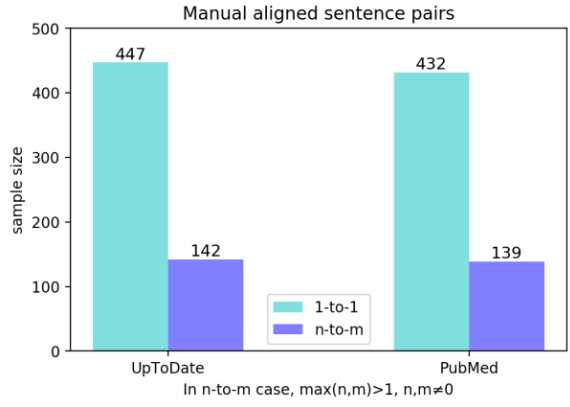

**Figure 3**

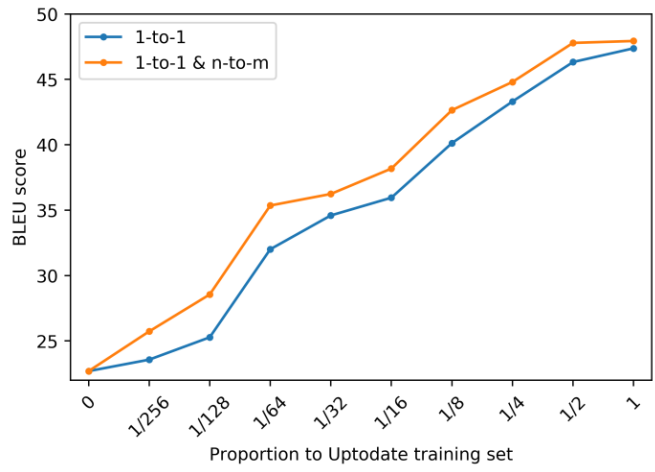

**Figure 4**

|  | Source | 给予 盐皮质激素 受体拮抗剂 (如， 螺内酯 或 依 普利 酮) 是 许多 HF 患者 标准 内科 治疗方案 的 一部分， 一项 关于 使用 罗格列酮 治疗 的 无 HF 的 糖尿病 患者 的 研究 显示， 螺内酯 比 袢 利尿剂 能 更 大程度 地 移除 液体 [ 69 ] 。 |
|---|---|---|
| Example 1 | Target | Administration of a mineralocorticoid receptor antagonist, such as spironolactone or eplerenone, is part of the standard medical regimen of many patients with HF and in a study of diabetic patients without HF who were treated with rosiglitazone, spironolactone produced more fluid removal than a loop diuretic [69]. |
|  | WMT18/ BioMed | The administration of a mineralocorticoid receptor antagonist (eg, spironolactone or eplerenone) is part of the standard medical regimen for many patients with HF, and in a study of patients with diabetes mellitus without HF who were treated with rosiglitazone, spironolactone provided greater fluid removal than a loop diuretic [69]. |
|  | WMT18/ None | Saline corticosteroid receptor antagonists were given as part of the standard internal therapy programme for many HF patients, a study of non-HF diabetic patients treated with roglitazone showed that LNE can remove liquids to a greater extent than cyclic diuretics. |

|  | Source | Thiazolidinediones can cause fluid retention. |
|---|---|---|
| Example 2 | Target | 噻唑烷 二 酮类 药物 可 导致 液体 潴留 。 |
|  | WMT18/ BioMed | 噻唑烷 二酮 类药物 可 引起 液体 潴留 |
|  | WMT18/ None | 硫磺 酰胺 能 引起 液体 滞留 。 |

|  | Source | Aegyptianellosis, eperythrozoonosis, grahamellosis and haemobartonellosis are diseases caused by procaryotic microorganisms which parasitize the erythrocytes of a variety of animals. |
|---|---|---|
| Example 3 | WMT18/ BioMed | 埃及伊蚊 、 红细胞增多症 、 钩端螺旋体病 和 血小板减少症 是 由 寄生 于 多种 动物 红细胞 的 原核生物 引起 的 疾病 。 |
|  | WMT18/ BioMed* | 埃及伊蚊 、 钩端螺旋体病 、 格雷 汉体 病 和 血 巴尔通体病 是 由 寄生 于 多种 动物 红细胞 的 原核生物 引起 的 疾病 。 |

**Figure 5**